\documentclass[runningheads]{llncs}

\usepackage{eccv}

\usepackage{eccvabbrv}
\usepackage{epsfig}
\usepackage{url}            %
\usepackage{booktabs}       %
\usepackage{amsfonts}       %
\usepackage{nicefrac}       %
\usepackage{microtype}      %
\usepackage{xcolor}         %
\usepackage{colortbl}         %
\usepackage{multirow}
\usepackage{graphicx}
\usepackage{amsmath}
\usepackage{amssymb}
\usepackage{bbm}
\usepackage{bm}
\usepackage{booktabs}
\usepackage{alias}
\usepackage{lib}
\usepackage{cuted}
\usepackage{xcolor}
\usepackage{cite}
\usepackage{array}
\newcolumntype{C}[1]{>{\centering\arraybackslash}p{#1}}
\usepackage{wrapfig}
\usepackage{subcaption}
\usepackage{pifont}%
\usepackage{setspace}
\newcommand{\cmark}{\ding{51}}%

\usepackage[accsupp]{axessibility}  %

\usepackage{hyperref}

\usepackage{orcidlink}

\begin{document}

\title{3D Reconstruction of Objects in Hands \\ without Real World 3D Supervision}

\author{Aditya Prakash \and Matthew Chang \and Matthew Jin \and Ruisen Tu \and Saurabh Gupta}

\authorrunning{A.~Prakash et al.}

\institute{University of Illinois Urbana-Champaign
\email{\{adityap9,mc48,mjin11,ruisent2,saurabhg\}@illinois.edu}\\
\url{https://bit.ly/WildHOI}}

\maketitle

\begin{abstract}
    Prior works for reconstructing hand-held objects from a single image train models on images paired with 3D shapes. Such data is challenging to gather in the real world at scale. Consequently, these approaches do not generalize well when presented with novel objects in in-the-wild settings. While 3D supervision is a major bottleneck, there is an abundance of a) in-the-wild raw video data showing hand-object interactions and b) synthetic 3D shape collections. In this paper, we propose modules to leverage 3D supervision from these sources to scale up the learning of models for reconstructing hand-held objects. Specifically, we extract multiview 2D mask supervision from videos and 3D shape priors from shape collections. We use these indirect 3D cues to train occupancy networks that predict the 3D shape of objects from a single RGB image. Our experiments in the challenging object generalization setting on in-the-wild MOW dataset show 11.6\% relative improvement over models trained with 3D supervision on existing datasets.
    \end{abstract}

\section{Introduction}
\seclabel{intro}

\begin{figure}
    \centering
    \includegraphics[width=\columnwidth,page=1]{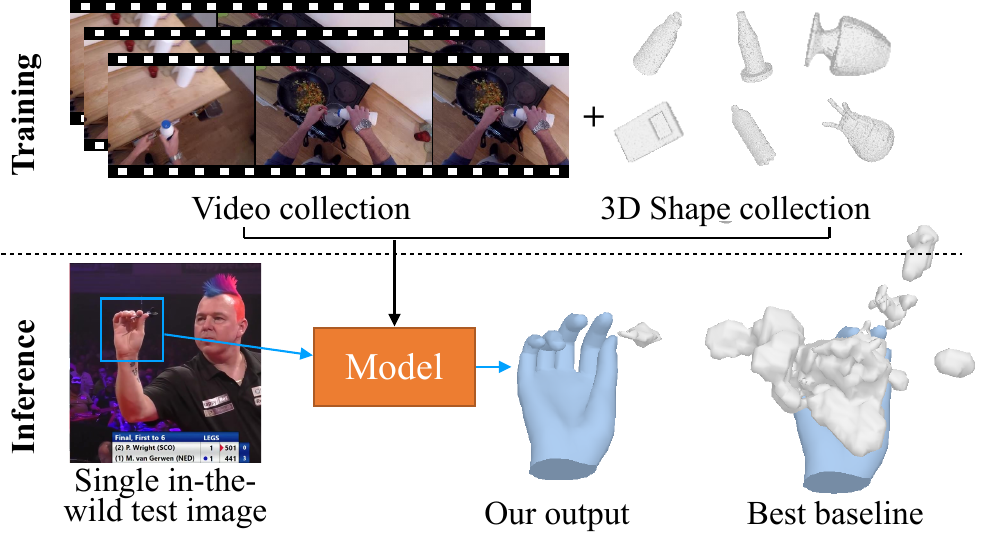}
    \caption{We propose modules to extract supervision from in-the-wild videos (\Secref{segm-sup}) \& learn shape priors from 3D object collections (\Secref{contact-sup}), to train occupancy networks which predict the 3D shapes of hand-held objects from a single image. This circumvents the need for paired real world 3D shape supervision used in existing works~\cite{ye2022ihoi,hasson19_obman}.}
    \figlabel{teaser}
\end{figure}

While 3D reconstruction of hand-held objects is important for AR/VR~\cite{Han2020TOG,Buckingham2021VR} and
robot learning applications~\cite{Mandikal2021CORL,Mandikal2021ICRA,Qin2022ECCV,Qin2022IROS,Ye2022ARXIV,Wu2022CORL}, {\it lack of 3D supervision outside of lab
settings} has made it challenging to produce models that work in the wild. This
paper develops techniques to improve the generalization capabilities of {\it
single image} hand-held object reconstruction methods by extracting supervision from in-the-wild videos \& synthetic shape collections showing hand-object interactions.

Collecting image datasets with ground truth 3D shapes for hand-held objects is
hard. Any visual scanning setups (via multiple \rgb/\rgbd cameras or
motion capture) require full visibility of the object which is not
available. Synthesizing realistic hand-object interaction is an open problem in
itself~\cite{Rijpkema1991SIGGRAPH,karunratanakul2020grasping,Jiang2021ICCV,Turpin2022ECCV}.
Manual alignment of template
shapes~\cite{cao2021reconstructing} is expensive, yet only approximate. Thus,
there is very little in-the-wild real-world data with ground truth 3D shapes
for hand-held objects. And while many past works have designed expressive
models to predict shapes of hand-held objects~\cite{ye2022ihoi,
hasson19_obman,karunratanakul2020grasping}, they are all held back due to the
limited amount of real-world 3D data available for training and suffer from
unsatisfactory performance on novel objects encountered in the wild.

While in-the-wild images with paired 3D shapes are rare, there are a) plenty of
in-the-wild videos containing multiple views of hand-held
objects~\cite{Damen2018ScalingEV, grauman2022ego4d} (\Figref{teaser}), b) large
catalogues of 3D object shapes~\cite{Chang2015ShapeNetAI} (\Figref{teaser}).
Shape collections provide 3D supervision but lack realistic hand grasps, videos
showcase realistic hand-object interaction but don't provide direct 3D
supervision. Either by itself seems insufficient, but can we combine
supervision from these diverse sources to improve generalization of single-image hand-held object reconstruction methods? 

Let's consider each cue one at a time.  While videos show multiple views of the
object, we unfortunately don't know the relative object pose in the different
views. Automatically extracting the object pose using structure from motion
techniques, \eg COLMAP~\cite{schoenberger2016sfm} doesn't work due to
insufficient number of feature matches on the object of interaction.  We
sidestep this problem by using {\it hand pose as a proxy for object pose}
(\Figref{sampled_points}). This is based on the observation that humans rarely
conduct in-hand manipulation in pick \& place tasks involving rigid objects. Thus, if we
assume that the hand and the object are rigidly moving together, then the
relative 6 DoF pose of the hand between pairs of frames reveals the relative 6
DoF pose of the object.  This reduces the SfM problem to an easier setting
where the motion is known.  Specifically, we use off-the-shelf FrankMocap
system~\cite{rong2020frankmocap} to obtain 6 DoF pose for the hand and
consequently the object's.  We then use our proposed 2D mask guided 3D sampling
module (\Secref{segm-sup}) to generate 3D supervision for the object shape
using object segmentation masks (\Figref{sampled_points}). This lets us train
on objects from 144 different categories, where as most methods currently train
on only a handful of categories ($< 20$).

While this works well for unoccluded parts of the object, this does not
generate reliable supervision for parts of the object that are occluded by the
hand (\Figref{teaser}). This brings us to the 3D shape catalogues, which we use
to extract shape priors.  This enables the model to learn to output contiguous
shapes even when the object is interrupted by the hand in the image, \eg it can
hallucinate a handle for a jug even when it is covered by the hand, because
jugs typically have one.  We adopt an adversarial training
framework~\cite{goodfellow2014generative} to train a discriminator to
differentiate between real shapes (from \obman~\cite{hasson19_obman}) and shapes predicted from the
model (\Figref{discriminator}). Unlike prior works~\cite{wu2016learning} which
train the discriminator on 3D inputs, we instead propose a 2D slice-based 3D
discriminator (\Secref{contact-sup}), which is computationally efficient and
learns better fine-grained shape information.

Our overall framework consists of an occupancy
network~\cite{mescheder2019occupancy} that predicts the 3D shape of hand-held
objects from a single image. We train this model on sequences curated from the
\visor dataset~\cite{darkhalil2022visor} and use the Obman
dataset~\cite{hasson19_obman} to build the shape prior. Training on diverse
real world data outside of lab settings, enabled by our innovations, 
leads our model (\name) to good generalization performance.
\name outperforms previous state-of-the-art models by 11.6\% in the challenging object generalization setting on \mow~\cite{cao2021reconstructing}. 

\section{Related Work}
\seclabel{related}

\boldparagraph{Reconstructing objects in hands}
Several works~\cite{hasson19_obman,ye2022ihoi,Chen2022ECCV,Chen2023CVPR,Zhang2023NEURIPS,karunratanakul2020grasping} have trained expressive architectures for predicting 3D shape from a single image using paired real world 3D supervision. Fitting object templates~\cite{cao2021reconstructing,Hasson2020LeveragingPC} or learned 3D shapes~\cite{Ye2023ICCV,Ye2024CVPR,fan2023hold,huang2022reconstructing} to videos using appearance cues~\cite{huang2022reconstructing,fan2023hold,cao2021reconstructing,Hasson2020LeveragingPC} or geometric priors~\cite{Ye2023ICCV,Ye2024CVPR} have also been explored. 
The most relevant work to ours is~\cite{ye2022ihoi}, which uses paired 3D supervision from synthetic~\cite{hasson19_obman} and small-scale real-world datasets to predict 3D shape from a single image. However, it does not generalize to novel object categories in the wild due to limited 3D supervision. Instead, we train our model on diverse object categories from in-the-wild videos by extracting multiview 2D supervision and learning shape priors from existing datasets, without any real-world 3D supervision. Note that our setting involves a single image input at test time and we use in-the-wild videos for training only. %

\boldparagraph{Hand-Object datasets with 3D object models}
Existing real-world hand-object datasets with 3D annotations are captured in lab settings and contain limited variation in objects, \eg HO3D~\cite{hampali2020honnotate}:10, H2O~\cite{Kwon2021ICCV}:8, FPHA~\cite{Hernando2018CVPR}:4, FreiHAND~\cite{Zimmermann2019ICCV}:35, ContactDB~\cite{Brahmbhatt2019CVPR}:50, ContactPose~\cite{Brahmbhatt2020ContactPoseAD}:25, DexYCB~\cite{Chao2021CVPR}:20, GRAB~\cite{taheri2020grab}: 51, \hofd~\cite{Liu2022CVPR}: 16 object categories. Collecting datasets with ground truth 3D shapes is difficult to scale since it often requires visual scanning setups (multiple cameras or motion capture). Synthesising realistic hand-object interaction is an open problem in itself~\cite{Rijpkema1991SIGGRAPH,karunratanakul2020grasping,Jiang2021ICCV,Turpin2022ECCV}. In this work, we curate sequences from in-the-wild \visor dataset containing 144 object categories and design modules to extract supervision for training occupancy networks. The closest to ours is MOW with 120 objects that we only use to test models to assess generalization.

\boldparagraph{Hand-Object Interactions in the wild}
There is a growing interest in understanding hands and how they interact with objects around them. Researchers have collected datasets~\cite{hasson19_obman,hampali2020honnotate,taheri2020grab,Chao2021CVPR,Kwon2021ICCV,Hampali2022CVPR,Liu2022CVPR} and trained models for detecting \& segmenting hands and associated objects of interaction~\cite{shan2020understanding, darkhalil2022visor,Tschernezki20213DV,Tschernezki20223DV}. Recognizing what hands are doing in images~\cite{Prakash2024Hands,chang2023look,zhu2023get} is also relevant: through grasp classification~\cite{karunratanakul2020grasping}, 2D pose estimation~\cite{rogez20143d, zimmermann2017learning}, and more recently 3D shape and pose estimation~\cite{romero2017embodied, hasson19_obman,rong2020frankmocap,ye2022ihoi,Tschernezki2023NEURIPS,Hasson2020LeveragingPC} for both hands and objects in contact.

\noindent \textbf{3D from single image without direct 3D supervision.}
Several works relax the need for direct 3D supervision by incorporating auxiliary shape cues during training, \eg multi-view consistency in masks~\cite{tulsiani2018multiview}, depth from single image~\cite{zhou2017unsupervised,Irshad2022ECCV,Lunayach2023ARXIV} or stereo~\cite{Hepper2023CVPR}, appearance~\cite{Yu2021pixelNeRFNR,choi2024handnerf,Truong2023CVPR,Irshad2023ICCV}. These have been applied to reconstruction of category specific~\cite{kar2015category,Lunayach2023ARXIV,Irshad2023ICCV,kanazawa2018learning} as well as generic objects~\cite{ye2021shelf,Yu2021pixelNeRFNR,choi2024handnerf}. However, directly applying these approaches to hand-held objects in the wild poses several challenges, \eg unknown camera, novel object categories, heavy occlusion, inaccurate depth estimates. 
In this work, we propose modules to extract supervision from in-the-wild videos using object masks~\cite{darkhalil2022visor} \& hand pose~\cite{rong2020frankmocap} and learn priors from synthetic collections of hand-held objects~\cite{hasson19_obman}.

\newcommand{\Real}[0]{\mathbb{R}}

\section{Approach}
\seclabel{approach}

We propose a novel framework for training 3D shape predictors from a single image without using any real world 3D supervision. Following prior work~\cite{ye2022ihoi}, we use implicit shape representation~\cite{mescheder2019occupancy,park2019deepsdf} for 3D objects.

\subsection{Preliminaries}
\seclabel{prelim}
Consider the recent \ihoi model for this task from Ye \etal~\cite{ye2022ihoi}. Given an input \rgb image, \ihoi uses a neural network to predict the SDF of 3D points. The prediction is done in the hand coordinate frame obtained using FrankMocap~\cite{rong2020frankmocap}, which outputs (a) hand articulation parameters $\thetaa$ (45 dimensional MANO hand pose~\cite{romero2010hands}), (b) global rotation $\thetaw$ of the wrist joint \wrt camera, (c) weak perspective camera $\theta^c$, with scale factor $s$ \& 2D translation ($t_x, t_y$), which is converted into a full perspective camera $K$. These can be used to project a 3D point $\bx$ into the image ($f$ is the focal length) as $\bx_p = K [T_{\thetaw}\bx + (t_x, t_y, f/s)]$

\definecolor{darkgreen}{rgb}{0,0.7,0}
\begin{figure}[t]
    \centering
    \includegraphics[width=\columnwidth]{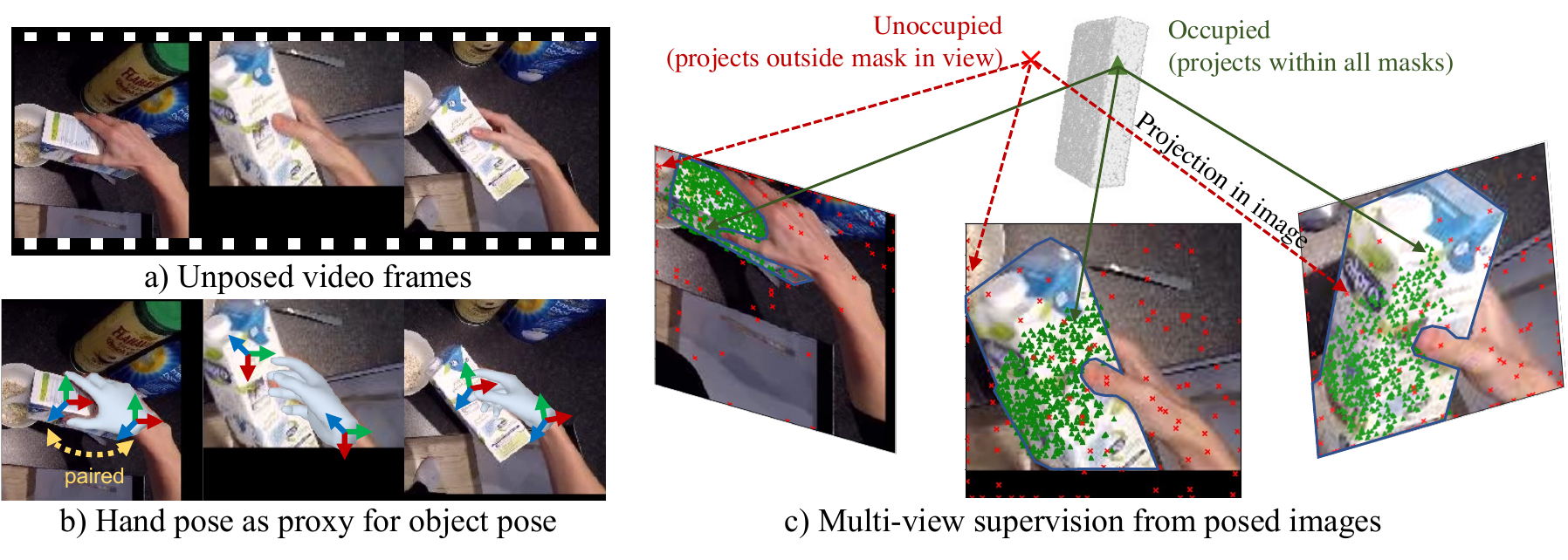}
    \caption{{\bf Registering objects via hand pose and 2D Mask guided 3D sampling.} (a) Consider unposed frames from in-the-wild videos. (b) We use hand pose from FrankMocap~\cite{rong2020frankmocap} as a proxy for object pose, thereby registering the different views. (c) We then use 2D object masks for labeling 3D points with occupancy (\Secref{segm-sup}). 3D points that project into the object mask in all views are considered as occupied ({\color{darkgreen}{green triangles}}), all other points are considered unoccupied ({\color{red}{red crosses}}). 
    (3D object in the figure is for visualization only, not used for sampling.)}
    \figlabel{sampled_points}
\end{figure}

Given a 3D point $\bx$ \& image $I$, \ihoi conditions the SDF prediction on: (a) global image features from a \resnet{50}~\cite{He2016DeepRL}, (b) pixel-aligned features~\cite{Saito2019PIFuPI} from intermediate layers of \resnet{50} at the projection $\bfx_p$ of $\bx$ in the image, (c) hand articulation features obtained by representing $\bx$ in the coordinate frame of 15 hand joints. 
This is realized as, ${\bf s} = \mathcal{F}(\bx; I, \theta, K)$.
Training $\mathcal{F}$ requires sampling 3D points $x$ around the object and corresponding SDF values $s$, $\theta = (\thetaa, \thetaw, \theta^c, K)$ are estimated from FrankMocap.

\subsection{2D Mask Guided 3D Sampling}
\seclabel{segm-sup}

Training models with implicit shape representation require supervision in the form of occupancy~\cite{mescheder2019occupancy} or SDF~\cite{park2019deepsdf} for 3D points sampled inside and outside the object. Note that the balanced sampling of points inside and outside the object is an important consideration for training good predictors. While existing approaches~\cite{ye2022ihoi,karunratanakul2020grasping,hasson19_obman} on this task use datasets with paired 3D supervision (3D object shape corresponding to 2D image), we operate in in-the-wild settings which do not contain 3D supervision. Instead, we propose a 2D mask guided 3D sampling strategy to obtain occupancy labels for training.

Consider multiple views $\{\I{I}{1}, \ldots, \I{I}{n}\}$ of a hand-held object (\Figref{sampled_points}), along with their masks $\{\I{M}{1}, \ldots, \I{M}{n}\}$. We can sample points $\bx$ in 3D space and project them into different views. Any point $x$ which projects into the object mask in all views is consider as occupied whereas if it projects outside the mask in even one of the views, it is considered as unoccupied. Thus, we get occupancy labels for a point $\bx$ as ${\bf{s}}^{gt} = \cap_{i=1}^{n} M_i^{\bx_{\I{p}{i}}}$.
Here, $M_i^{\bx_{\I{p}{i}}}$=$1$ if $x_{p_{i}}$ lies inside the mask $M_{i}$ \& $0$ otherwise. Note that it is not possible to obtain SDF values in this manner, since distance to the object surface cannot be estimated in the absence of 3D objects models. While we can obtain 3D occupancy labels using this strategy, there are two important considerations: camera poses are unknown (required for projection) \& how to balance the sampling of points inside \& outside the object.

\boldparagraph{Camera pose}
We assume that the hand is rigidly moving with the object. This is not an unreasonable assumption, as humans rarely do in-hand manipulation in pick \& place tasks involving small rigid objects. Thus, the relative pose of hand between different views reveals the relative pose of the object. This lets use the hand pose predicted by FrankMocap $\{\I{\theta}{1}, \ldots, \I{\theta}{n}\}$ to register the different views.

\boldparagraph{Balanced sampling}
In the absence of 3D object models, a natural choice is to sample points uniformly in 3D space. However, this leads to most points lying outside the object because the object location is unknown. Instead, we sample points in the hand coordinate frame.
Consider the total number of points to be $q$. We adopt several strategies for balanced sampling for points inside ($s^{gt}=1$) and outside the object ($s^{gt}=0$). We uniformly sample $q/2$ 3D points $\bx \in \Real^3$ in the normalized hand coordinate frame and project these into all the available views. Since all these $q/2$ points may not be occupied, we use rejection sampling to repeat the procedure, for maximum of $t=50$ times or until we get $q/2$ occupied points. Also, all points projecting into the hand mask in all views and vertices of the MANO~\cite{romero2017embodied} hand are labeled as unoccupied.

Formally, for images $\{\I{I}{1}, \ldots, \I{I}{n}\}$ with object masks $\{\I{M}{1}, \ldots, \I{M}{n}\}$, hand masks $\{\I{H}{1}, \ldots, \I{H}{n}\}$ and MANO vertices $\{\I{V}{1}, \ldots, \I{V}{n}\}$, ${\bf{s}}^{gt}$ for $\bx$ is:
\begin{equation}
    {\bf{s}}^{gt} = \begin{cases}
    1 & \text{if } \cap_{i=1}^{n} M_i^{\bx_{\I{p}{i}}} \text{ and } \cap_{i=1}^{n} \neg H_i^{\bx_{\I{p}{i}}} \text{ and } \cup_{i=1}^{n} \neg V_i^{\bx}\\
    0 & \text{otherwise}
    \end{cases}
\end{equation}
where $\bx_{\I{p}{i}}$ is the projection of $\bx$, $M_i^{\bx_{\I{p}{i}}}$=$1$ if $x_{p_{i}}$ lies inside $M_i$, $H_i^{\bx_{\I{p}{i}}}$=$1$ if $x_{p_{i}}$ lies inside $H_i$, $V_i^{\bx}$$=1$ if $\bx$ belongs to $V_i$ and $\neg$ is the logical negation operator. 

Note that, due to hand occlusions and errors in FrankMocap predictions, it is possible that some 3D points belonging to the object are not projected into the object masks but we do not want to label these points as unoccupied. So we disregard points which project onto the object mask in some views and hand mask in other views as these points could belong to object due to hand occlusion.

This is reminiscent of the visual hull algorithm~\cite{laurentini1994visual, matusik2000image}, which generates 3D reconstruction by carving out space that projects outside the segmentation in any view. Visual hull algorithms need multiple views at test time to generate any output. In contrast, we are doing this at training time to obtain supervision for $\mathcal{F}(\bx; \I{I}{1}, \I{\theta}{1}, \I{K}{1})$, which makes predictions from a single view.

\boldparagraph{Training}
We use cross-entropy loss (CE) to train $\mathcal{F}$ using ground truth ${\bfs}^{gt}$: %
\begin{equation}
    \mathcal{L}_\text{visual-hull} = \text{CE}(\mathcal{F(\bx)}, {\bfs}^{gt})
\end{equation}

To further regularize training, we also encourage the occupancy prediction from different views to be consistent with each other. Since our predictions are already in the hand coordinate frame, which is common across all views, this can be done by minimizing $\mathcal{L}_\text{consistency}$ for different views $i$ \& $j$ of the same object.
\begin{equation}
    \mathcal{L}_\text{consistency}= \sum_{\bfx \in \Real^3,i \ne j} \text{CE}\left(
        \mathcal{F}(\bx; \I{I}{i}, \I{\theta}{i}, \I{K}{i}), \mathcal{F}(\bx; \I{I}{j}, \I{\theta}{j}, \I{K}{j})\right)
\end{equation}

\begin{figure}[t]
    \centering
    \includegraphics[width=0.8\columnwidth]{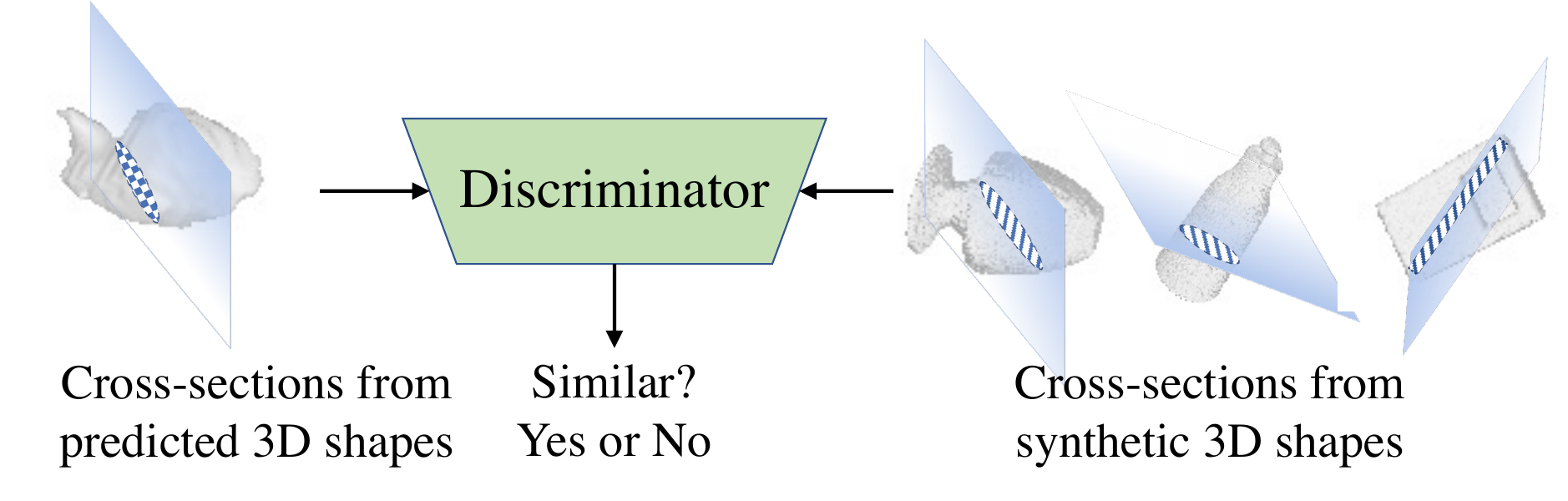}
    \caption{{\bf 2D slice based 3D discriminator}. We learn data-driven 3D shape priors using hand-held objects from \obman dataset. We sample planes through the object (shown above in blue), resulting in a 2D cross-section map. We pass occupancy predictions on points from these cross-sections through a discriminator which tries to distinguish cross-sections of predicted 3D shapes from cross-sections of \obman objects (\Secref{contact-sup}).}
    \figlabel{discriminator}
\end{figure}

\subsection{2D Slice based 3D Discriminator as Shape Prior}
\seclabel{contact-sup}
We adopt an adversarial training framework~\cite{goodfellow2014generative} to build a prior on shapes of hand-held objects and use it to supervise the training of the occupancy prediction function $\mathcal{F}(\bx; \I{I}{1}, \I{\thetaa}{1}, \I{\thetaw}{1}, \I{K}{1})$. As such a prior can be challenging to hand-craft, we build it in a data-driven way.
We use 3D shape repository from synthetic datasets~\cite{hasson19_obman}, which contain more than 2.5K hand-held objects, to learn the prior. Specifically, we train a discriminator $\mathcal{D}$ to differentiate between 3D shapes from \obman~\cite{hasson19_obman} and generated shapes as predicted by $\mathcal{F}$. We derive supervision for $\mathcal{F}$ by encouraging it to predict shapes that are real as per $\mathcal{D}$.

A natural choice is to train the discriminator with 3D input, \eg $N \times N \times N$ cube in 3D voxel space~\cite{wu2016learning}. One way to do this is to sample $N^3$ 3D points in the hand coordinate frame and run a forward pass through $\mathcal{F}$ to get the occupancy for each of these points. However this is computationally expensive and often leads to large imbalance as most points lie outside the object (we ablate this in~\Secref{ablation}). Instead, we propose a novel 2D slice based 3D discriminator which operates on arbitrary 2D slices. There are computed by taking the cross-section of 2D planes with 3D shapes and sampling 3D points that lie on these 2D cross-sections. The key intuition here is that the discriminator sees {\it different randomly sampled} 2D slides during the course of training, which helps it to learn fine-grained shape information. \Eg for a sphere, all cross-sections are circular but for a cylinder, most are oval. This helps distinguish between different 3D shapes.

\boldparagraph{Sampling 2D slices}
There are several important considerations in sampling 2D slices. First, uniformly sampling 2D planes often leads to most points lying outside the object, which is not useful for training the discriminator. Instead, we sample 2D planes that pass through the origin in the hand coordinate system. Since the objects are in contact with the hand, the sampled points are more likely to encompass the object. Then, we rotate the sampled 2D planes by arbitrary angles so that they are not axis aligned to better capture fine-grained shape information. We ablate all these design choices in~\Secref{ablation}. This sampling function $\mathcal{Z}$ results in a set of 2D planes on which 3D points are uniformly sampled.

\boldparagraph{Training}
We pass the sampled points from 2D slices of the generated 3D shape through $\mathcal{F}$ to get the corresponding occupancy values $S^{\text{gen}}$. This represents the generated 3D shape. We adopt the same strategy for representing 3D shapes from \obman (used as real shapes) but use the predictions $S^{\text{real}}$ of the occupancy network overfitted on \obman. As they come from a overfitted model, they generally match the ground truth slices well but at the same time are soft and prevent the discriminator from cheating. 

We train the discriminator $\mathcal{D}$ to differentiate between $S^{\text{gen}}$ \& $S^{\text{real}}$ using the least squares formulation~\cite{Mao2017ICCV} for discriminator loss. We derive supervision for $\mathcal{F}$ by computing gradients through $\mathcal{D}$ on the occupancy values at the sampled points to maximize the realism of the generated shapes.
\begin{align}
    \mathcal{L}_\text{adv}^\mathcal{D} &= [ \mathcal{D}(S^{\text{real}})-1 ]^2 + [ \mathcal{D}(S^{\text{gen}}) ]^2 \nonumber \\
    \mathcal{L}_\text{adv}^\mathcal{F} &= [ \mathcal{D}(S^{\text{gen}})-1 ]^2 \nonumber \\
    \mathcal{L}_\text{shape-prior} &= \lambda_f \mathcal{L}_\text{adv}(\mathcal{F}) + \lambda_d \mathcal{L}_\text{adv}(\mathcal{D})
\end{align}

\subsection{Training Details}
\seclabel{loss}
We train $\mathcal{F}$ \& $\mathcal{D}$ in an alternating manner with 2 iterations of $F$ for every iteration of $D$. The total loss for training our framework is:
\begin{align}
    \mathcal{L_{\mathcal{F}}} &= \lambda_v \mathcal{L}_\text{visual-hull} + \lambda_c \mathcal{L}_\text{consistency} + \lambda_f \mathcal{L}_\text{adv}^\mathcal{F} \nonumber \\
    \mathcal{L_{\mathcal{D}}} &= \lambda_d \mathcal{L}_\text{adv}^\mathcal{D}
\end{align}
Following standard practice~\cite{ye2022ihoi}, we pretrain on synthetic \obman. We train our model jointly on \obman (3D supervision, shape priors) \& \visor (2D supervision) with a dataset ratio of \obman:\visor as 1:2. We use batch size of 64, learning rate of 1e-5 across 4 NVIDIA A40 GPUs \& loss weights as $\lambda_v=1, \lambda_c=1$, $\lambda_f=0.25$, $\lambda_d=0.25$. Please refer to supplementary for more details.

\begin{figure}[t]
    \centering
    \includegraphics[width=\columnwidth,page=3]{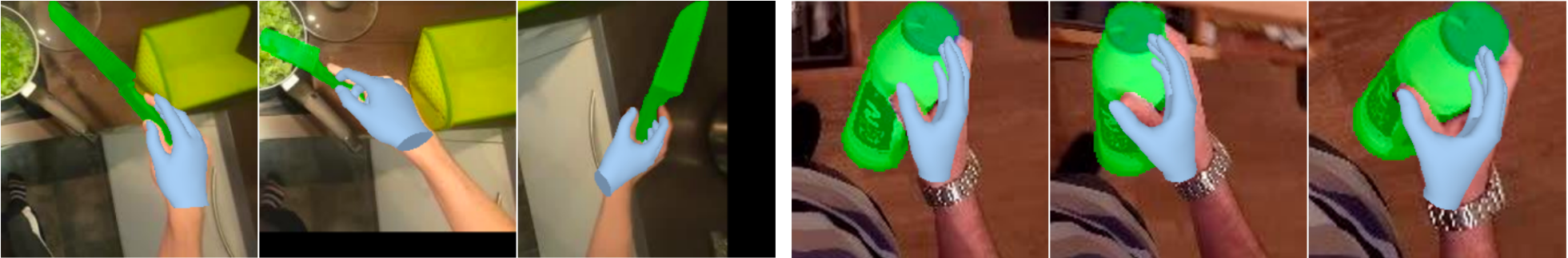}
    \caption{{\bf VISOR visualizations}. Using existing hand pose estimation techniques~\cite{rong2020frankmocap}, we are able to track the objects in relation to hands through time in in-the-wild videos. We visualize these tracks along with object masks from the VISOR dataset~\cite{darkhalil2022visor}. This form of data, where objects move rigidly relative to hands, is used to train our model to learn 3D shape of hand-held objects.}
    \figlabel{visor_hands}
\end{figure}

\subsection{Constructing \visorhoi Dataset}
\seclabel{visor-prep}

Our framework requires dataset containing multi-view images of rigid hand-object interactions in the wild, with 3D hand pose and 2D object masks. To construct such a dataset, we consider \visor~\cite{darkhalil2022visor} which provides 2D tracks for hands, objects they are interacting with and their segmentation masks. It contains a rich set of hand-object interactions, \eg taking out milk from the fridge, pouring oil from bottles, kneading dough, cutting vegetables, and stirring noodles in a wok. Our interest is in the 3D reconstruction of {\it rigid} objects which are {\it in-contact} with a hand, but there are no 3D object annotations in VISOR. Hence, we process it to prepare a dataset for training our model.

We first sample a subset of VISOR involving hand-object contact, using available contact annotations. We select object tracks where only one hand is in consistent contact with the object. This leaves us with 14768 object tracks from the original \visor dataset. We then manually filter this subset to select a subset that showcases manipulation of rigid objects with a single hand. This leaves us with 604 video snippets showing hands interacting with different objects.

\boldparagraph{Processing hands on VISOR}
We rely on the 3D hand poses to set up the output coordinate frame, compute hand articulation features, and more importantly to register the different frames together~\cite{ma2022virtual, tzionas20153d}. These hand poses are estimated using FrankMocap, which may not always be accurate. To remove erroneous poses, we employ automated filtering using the uncertainty estimate technique from Bahat \& Shakhnarovich~\cite{bahat2018confidence} following 3D human pose literature~\cite{rockwell2020full}. 
Specifically, we obtain 3D hand pose predictions on five different versions of the image, augmented by different fixed translations 
. The uncertainty estimate for a given image is computed as the standard deviation of reprojection locations of MANO vertices across these 5 image versions. This sidesteps the need to hand-specify the trade-off between translation, rotation, and articulation parameters that are part of the 3D hand pose output. This leaves us with 473 video snippets consisting of 144 object categories. This object diversity is 4$\times$ larger than existing datasets~\cite{Yang2022CVPR,hampali2020honnotate,Hampali2022CVPR,Liu2022CVPR,Kwon2021ICCV} used for our task, typically containing 10 to 32 object categories. We refer to this dataset as~\visorhoi, some example object sequences are shown in \Figref{visor_hands}. Note the {\it incidental} multiple views and relative consistency in hand and object pose over the course of interaction.

\section{Experiments}
\seclabel{expts}

\begin{table*}[t]
    \centering
    \renewcommand{\arraystretch}{1.05}
    \caption{{\bf Generalization to novel objects in the wild.} We report F-score at 5mm \& 10 mm, Chamfer distance (CD, mm) for object generalization splits on \mow. We compare with AC-OCC \& AC-SDF trained on different combinations of datasets with full 3D supervision. Our approach outperforms baselines across all metrics without using real-world 3D supervision (Relative \% improvement \wrt best baseline in {\color{darkgreen}{green}}).}
    \tablelabel{obj_gen}
    \setlength{\tabcolsep}{4pt}
    \resizebox{\linewidth}{!}{
    \begin{tabular}{l f ccc}
        \toprule
        \bf Method      & \bf Dataset and supervision used & \bf F@5 $\uparrow$ & \bf F@10 $\uparrow$ & \bf CD $\downarrow$ \\
        \midrule
        AC-OCC & \obman (Synthetic 3D)                              & 0.095     & 0.179     & 8.69 \\
        AC-SDF~\cite{ye2022ihoi} & \obman (Synthetic 3D)                              & 0.108     & 0.199     & 7.82 \\
        AC-SDF~\cite{ye2022ihoi} & \obman (Synthetic 3D) + \hotd (Lab 3D)             & 0.082     & 0.159     & 7.52 \\
        AC-SDF~\cite{ye2022ihoi} & \obman (Synthetic 3D) + \hotd (Lab 3D) + \hofd (3D)& 0.095     & 0.193     & 7.43 \\
        \name (Ours) & \obman (Synthetic 3D) + \visor (2D Masks) + Shape priors & {\bf 0.121}{\gain{darkgreen}{+10.7}} & {\bf 0.220}{\gain{darkgreen}{+10.6}} & {\bf 6.76}{\gain{darkgreen}{+13.5}} \\
        \bottomrule
    \end{tabular}}
\end{table*}

\subsection{Protocols}
We use 4 datasets for training (\obman~\cite{hasson19_obman}, \visor~\cite{darkhalil2022visor}, \hotd~\cite{hampali2020honnotate}, \hofd~\cite{Liu2022CVPR}) and 2 datasets (\mow~\cite{cao2021reconstructing}, \hotd) for evaluation. Different methods are trained on different datasets, depending on the specific evaluation setting.

\boldparagraph{Training datasets}
\obman is a large scale synthetic hand-object dataset with 2.5K objects and 3D supervision. \hotd \& \hofd are real world datasets collected in lab settings with 3D annotations. \hotd contains 10 YCB~\cite{alli2015TheYO} objects whereas \hofd contains 16 object categories, out of which 7 are rigid. \visor does not contain any 3D supervision. Instead, we use the process described in~\Secref{visor-prep}, to extract supervision from \visor, resulting in 144 object categories.

The baselines are trained with different combinations of \hotd \& \hofd~\cite{Liu2022CVPR}. As our method does not require 3D ground truth, we do not use these datasets for training. Instead, we use auxiliary supervision from \visorhoi (\Secref{visor-prep}) \& learn shape priors using \obman. \visor does not have 3D annotations and can not be used to train the baselines. Note that all models are initialized from the model pretrained on \obman for fair comparisons, following protocol~\cite{ye2022ihoi}.

\boldparagraph {Evaluation datasets}
We focus on the challenging zero-shot generalization to novel objects in-the-wild setting. We use \mow~\cite{cao2021reconstructing} dataset which contains images from YouTube, spanning 120 object templates. Note that these types of images have not been seen during training. To be consistent with prior work~\cite{ye2022ihoi}, we also use \hotd for evaluation, consisting of 1221 testing images across 10 objects. While~\cite{ye2022ihoi} operate in view generalization setting, \ie, making predictions on novel views of training objects, we also consider the more challenging object generalization setting. Almost all of our experiments are conducted in the {\it object generalization setting} where we assess predictions on novel objects across datasets.

\begin{figure}[t]
\begin{minipage}[t]{0.50\linewidth}
    {
    \renewcommand{\arraystretch}{1.1}
    \captionof{table}{{\bf \hotd Object generalization.} We outperform AC-OCC \& AC-SDF trained on different datasets with 3D supervision.}
    \tablelabel{obj_gen_ho3d}
    \setlength{\tabcolsep}{3pt}
    \centering
    \resizebox{\linewidth}{!}{
    \begin{tabular}{llccc}
        \toprule
        \bf Method & \bf Supervision (\obman +) & \bf F@5 & \bf F@10 & \bf CD \\
        \midrule
        AC-OCC & - & 0.18 & 0.33 & 4.39 \\
        AC-SDF & - & 0.17 & 0.33 & 3.72 \\
        AC-SDF & \mow (3D) & 0.17 & 0.33 & 3.84 \\
        AC-SDF & \mow (3D) + \hofd (3D) & 0.17 & 0.33 & 3.63 \\
        Ours & \visor (Multi-view 2D) & \bf 0.20 & \bf 0.35 & \bf 3.39 \\
        \bottomrule
    \end{tabular}}
    }
\end{minipage}
\hfill
\begin{minipage}[t]{0.48\linewidth}
    {
    \renewcommand{\arraystretch}{1.1}
    \setlength{\tabcolsep}{3pt}
    \centering
    \captionof{table}{{\bf \hotd View generalization.} We outperform HO~\cite{hasson19_obman} \& GF~\cite{karunratanakul2020grasping}, trained on \hotd with full 3D supervision.}
    \tablelabel{view_gen_ho3d}
    \resizebox{\linewidth}{!}{
    \begin{tabular}{llccc}
        \toprule
        \bf Method & \bf Supervision (\obman +) & \bf F@5 & \bf F@10 & \bf CD \\
        \midrule
        AC-SDF & -                           & 0.17 & 0.32 & 3.72 \\
        HO~\cite{hasson19_obman} & \hotd (3D)             & 0.11 & 0.22 & 4.19 \\
        GF~\cite{karunratanakul2020grasping} & \hotd (3D) & 0.12 & 0.24 & 4.96 \\
        Ours & \hotd (Multi-view 2D)                   & \bf 0.23 & \bf 0.43 & \bf 1.41 \\
        \bottomrule
    \end{tabular}}
    }
\end{minipage}
\end{figure}

\boldparagraph{Metrics}
Following~\cite{ye2022ihoi,tatarchenko2019single}, we report Chamfer distance (CD) and F-score at 5mm \& 10mm thresholds. F-score evaluates the distance between object surfaces as the harmonic mean between precision \& recall. Precision measures accuracy of the reconstruction as \% of reconstructed points that lie within a certain distance to ground truth. Recall measures completeness of the reconstruction as \% of points, on the ground truth, that lie within a certain distance to the reconstruction. CD computes sum of distances for each pair of nearest neighbors in the two point clouds. We report mean CD \& F-score over all test objects.

\boldparagraph{Baselines} We compare our model with \ihoi trained in supervised manner using 3D ground truth on different combination of datasets in different settings: (1) For object generalization on \mow in the wild, \ihoi is trained on \obman, \obman + \hotd, \obman + \hotd + \hofd, (2) For object generalization on \hotd, \ihoi is trained on \obman, \obman + \mow, \obman + \mow + \hofd, (3) For view generalization on \hotd, \ihoi is trained on \obman + \hotd. We also compare with an occupancy variant of AC-SDF (AC-OCC) and recent published methods with different forms of SDF representation, \eg AlignSDF~\cite{Chen2022ECCV}, gSDF~\cite{Chen2023CVPR}, DDFHO~\cite{Zhang2023NEURIPS}. Note that the \visor dataset cannot be used for training since it does not have 3D supervision. For the view generalization setting on \hotd, we also compare with HO~\cite{hasson19_obman} \& GF~\cite{karunratanakul2020grasping} trained with 3D ground truth on \obman + \hotd.
Recent works~\cite{Niemeyer2020CVPR,Yariv2020NEURIPS} on unsupervised reconstruction of objects require several views or depth, which are not available in our setting.

\subsection{Results}
\seclabel{results}

\begin{table}[t]
    \begin{minipage}[t]{0.48\linewidth}
    \centering
    \caption{\textbf{Comparison with relevant methods.} Our approach also outperforms gSDF, AlignSDF \& DDFHO (trained in the same setting as ours) in zero-shot generalization to \mow across most metrics.}
    \tablelabel{tab:relevant_methods}
    \small
    \resizebox{\linewidth}{!}
    {
    \setlength{\tabcolsep}{2.5pt}
    \begin{tabular}{l ccc}
      \toprule
      \bf Method & \bf F@5 $\uparrow$ & \bf F@10 $\uparrow$ & \bf CD $\downarrow$ \\
      \midrule
      AC-SDF~\cite{ye2022ihoi} & 0.108 & 0.199& 7.82 \\
      AlignSDF~\cite{Chen2022ECCV} & 0.099 & 0.182 & 8.30 \\
      gSDF~\cite{Chen2023CVPR} & 0.107 & 0.197 & 7.50 \\
      DDFHO~\cite{Zhang2023NEURIPS} & 0.094 & 0.166 & 3.06 \\
      HORSE (Ours) & 0.121 & 0.220 & 6.76 \\
      \bottomrule
    \end{tabular}
    }
    \end{minipage}
    \hfill
    \begin{minipage}[t]{0.48\linewidth}
    \centering
    \caption{\textbf{3D \vs 2D input to discriminator.} 
    Training with 3D inputs (at different resolutions) perform worse, likely due to coarse sampling resulting in very few points inside the object.
    }
    \tablelabel{disc_input}
    \small
    \resizebox{\linewidth}{!}
    {
    \setlength{\tabcolsep}{3.5pt}
    \begin{tabular}{l c c c}
    \toprule
    \bf Disc. input & \bf F@5 $\uparrow$ & \bf F@10 $\uparrow$ & \bf CD $\downarrow$\\
    \midrule
       No disc. & 0.117 & 0.216 & 6.93 \\
       $10\times 10\times 10$ & 0.120 & 0.218 & 7.29 \\
       $16\times 16\times 16$ & 0.115 & 0.209 & 7.79 \\
       $32\times 32\times 32$ & 0.104 & 0.191 & 7.83 \\
       2D slices & \bf 0.121 & \bf 0.220 & \bf 6.76 \\
    \bottomrule
    \end{tabular}
    }
    \end{minipage}
\end{table}

\boldparagraph{Object generalization in the wild}. We first examine if the auxiliary supervision from visual hull and shape prior is useful for generalization to novel objects in the wild. We evaluate on \mow in~\Tableref{obj_gen} and compare with AC-OCC \& AC-SDF trained on different combinations of \obman, \hotd, \hofd datasets with 3D supervision. Our approach provides gains of 24.3\% compared to AC-OCC (trained on \obman) and 11.6\% on AC-SDF (trained on \obman). This shows the benefits of our supervision cues in the wild over training on just large scale synthetic data with 3D supervision. We also outperform AC-SDF trained on \obman + \hotd + \hofd with full 3D supervision by 16.8\% across all metrics. This indicates that our supervision cues from in-the-wild \visor are better than using 3D supervision on lab datasets with limited diversity in objects. We also outperform relevant methods that use different forms of SDF representations, \eg AlignSDF, gSDF \& DDFHO across most metrics (\Tableref{tab:relevant_methods}). Note that our contributions are orthogonal and could be combined with these works.

\noindent {\bf Adding 3D supervision to AC-SDF}. In~\Tableref{obj_gen}, we observe that adding more data from \hotd \& \hofd to AC-SDF training did not help in zero-shot generalization to \mow. Instead, the performance drops compared to AC-SDF trained on \obman. This is likely due to limited diversity in \hotd: 10 YCB objects, \hofd: 7 rigid object categories \& the model overfitting to these categories.

\boldparagraph{Object generalization on \hotd} Our approach is better than AC-OCC \& AC-SDF trained on different datasets with 3D supervision (\Tableref{obj_gen_ho3d}). This further shows the benefits of auxiliary supervision from \visor for object generalization. Also, AC-SDF does not benefit from \mow \& \hofd. This could because \hotd evaluates on 10 objects only and they may not be present in \mow or \hofd.

\noindent {\bf Occupancy vs SDF}. We see that SDF formulation is better than occupancy when trained with full 3D supervision (AC-OCC \vs AC-SDF). In contrast, we find SDF training to be unstable (does not give meaningful predictions) with auxiliary supervision. This could be because regressing continuous SDF values with weak supervision is harder than binary classification for occupancy values.

\noindent {\bf View generalization results on \hotd}. In~\Tableref{view_gen_ho3d}, we see gains with using supervision cues over just training on synthetic data, consistent with trends in the object generalization setting. We also outperform HO~\cite{hasson19_obman} \& GF~\cite{karunratanakul2020grasping}, both trained on \hotd using full 3D supervision. We outperform these methods even without any images from \hotd (last row in \Tableref{obj_gen} \vs GF \& HO in \tableref{view_gen_ho3d}), likely due to use of more expressive pixel-aligned \& hand articulation features.

\subsection{Ablation Study}
\seclabel{ablation}

\newcommand{\fttext}[0]{\footnote{Note that FrankMocap was trained on \hotd dataset, thus FrankMocap predictions are already very good on \hotd.}}

\begin{table}[t]
    \begin{minipage}[t]{0.46\linewidth}
    \centering
    \caption{{\bf Supervision quality on \hotd.}
    Automated filtering to remove incorrect hand poses improves results \& using ground truth hand pose differs little compared to predicted pose.\textsuperscript{1}
    } %
    \tablelabel{ablation_ho3d}
    \resizebox{\linewidth}{!}{
    \begin{tabular}{llccc}
        \toprule
        & \bf F@5 $\uparrow$ & \bf F@10 $\uparrow$ & \bf CD $\downarrow$ \\
        \midrule
        \name (base setting)        & 0.234 & 0.434 & 1.41 \\
        \quad no training on \hotd  & 0.175 & 0.329 & 3.72 \\
        \quad w/o filtering         & 0.213 & 0.405 & 1.42 \\
        \quad w/ ground truth pose\footnotemark[1] & 0.243 & 0.444 & 1.39 \\
        \bottomrule
    \end{tabular}}
    \end{minipage}
    \hfill
    \begin{minipage}[t]{0.51\linewidth}
    {
    \setlength{\tabcolsep}{2.5pt}
    \centering
    \caption{{\bf Role of different loss functions.} We report F-score at 5mm \& 10mm, Chamfer distance (CD, mm) for different variants of our model on \mow. All losses are effective \& multiview supervision leads to largest gain.}
    \tablelabel{ablation}
    \resizebox{\linewidth}{!}{
    \begin{tabular}{ccccccc}
        \toprule
        \bf \Lobman & \bf \Locc & \bf \Lconsis & \bf \Lshape & \bf F@5 $\uparrow$ & \bf F@10 $\uparrow$ & \bf CD $\downarrow$ \\
        \midrule
        \cmark &        &        & & 0.095 & 0.181 & 8.69 \\
        \cmark & \cmark &        & & 0.111 & 0.205 & 7.26 \\
        \cmark &        & \cmark & & 0.073 & 0.132 & 12.75 \\
        \cmark &        & & \cmark & 0.097 & 0.175 & 10.29 \\
        \cmark & \cmark & \cmark & & 0.117 & 0.216 & 6.93 \\
        \cmark & \cmark & \cmark & \cmark & \bf 0.121 & \bf 0.220 & \bf 6.76\\
        \bottomrule
    \end{tabular}}
    }
    \end{minipage}
\end{table}

\noindent {\bf Analysis of supervision quality.} 
We also observe in \Tableref{view_gen_ho3d} that our method is able to bridge more than 40\% of the gap between no training on \hotd to training with full 3D supervision.
We further use the view generalization setting to assess the quality of 2D object mask supervision used in our method in~\Tableref{ablation_ho3d}. 
Our automated filtering of frames with inaccurate hand poses (as described in \Secref{visor-prep})  is crucial for good performance. Also, little is lost from using hand pose as a proxy for object pose on the \hotd dataset.
\footnote{\label{ftnote}While~\cite{ye2022ihoi} uses similar contrast between predicted \vs ground truth hands to make claims, we note that those claims \& this result should be taken with a grain of salt. FrankMocap is trained on \hotd, so its predictions on \hotd are better than they would be on unseen data. As most of our models are trained on \visor (not used for training FrankMocap), our other experiments do not suffer from this issue.}

\boldparagraph{Role of different loss terms}
We experiment with multiple variants of our model to assess the importance of different loss terms. We start with the AC-OCC model trained on \obman and gradually add \Locc, \Lconsis, and \Lshape. %
From the results in~\Tableref{ablation}, we observe that \Locc is more effective than \Lconsis and using them together provides further benefits. Moreover, \Lshape improves performance on top of \Lconsis and \Locc.

\begin{table}[t]
    \begin{minipage}[t]{0.46\linewidth}
    \centering
    \captionof{table}{\textbf{Design choices for mask guided sampling.} Uniformly sampling points is much worse than the rejection sampling used in our method. Using negative points from hand masks is useful.}
    \tablelabel{mask_sampling}
    \small
    \resizebox{\linewidth}{!}
    {
    \begin{tabular}{c c c c}
    \toprule
    Sampling method & F@5 $\uparrow$ & F@10 $\uparrow$ & CD $\downarrow$\\
    \midrule
       Uniform & 0.093 & 0.166 & 10.29 \\
       Ours (no hand points) & 0.113 & 0.207 & 7.69 \\
       Ours & \bf 0.117 & \bf 0.216 & \bf 6.93 \\
    \bottomrule
    \end{tabular}
    }
    \end{minipage}
    \hfill
    \begin{minipage}[t]{0.52\linewidth}
    \centering
    \caption{\textbf{Sampling method for 2D planes.} Sampling planes through origin of hand coordinate system \& rotated randomly performs the best compared to sampling axis-aligned planes either uniformly or through origin.}
    \tablelabel{disc_planes}
    \small
    \resizebox{\linewidth}{!}
    {
    \begin{tabular}{c c c c}
    \toprule
    Sampling method & F@5 $\uparrow$ & F@10 $\uparrow$ & CD $\downarrow$\\
    \midrule
       Uniform (axis-aligned) & 0.115 & 0.208 & 7.01 \\
       Origin (axis-aligned) & 0.098 & 0.183 & 8.52 \\
       Origin (random rotation) & \bf 0.121 & \bf 0.220 & \bf 6.76 \\
    \bottomrule
    \end{tabular}
    }
    \end{minipage}
\end{table}

\boldparagraph{3D vs 2D input to discriminator}
We also consider 3D volumes as input to the discriminator (instead of 2D cross-sections). For this, we need to sample 64x64x64 (=262144) points \& run several forward passes of our model to get occupancies
Since this is computationally expensive, we sample points at coarser resolutions: 32x32x32, 16x16x16, 10x10x10. We use 32x32 size 2D slices, so 10x10x10 3D volume has no. of points \& takes similar compute. We see that 2D slices perform better than 3D volumes (\Tableref{disc_input}). Also, the performance gets worse with increase in the sampled 3D volume, likely due to 3D sampling being so coarse that very few points lie inside the object, thus unable to capture fine-grained shape.

\boldparagraph{Sampling 2D slices for discriminator}
We ablate different design choices (\Secref{contact-sup}) in~\Tableref{disc_planes}. We observe that sampling 2D planes through origin of the hand coordinate system and rotated randomly performs the best compared to sampling axis-aligned frames either uniformly or through origin.

\boldparagraph{Design choices for mask guided sampling}
We run rejection sampling (with hand \& object masks) to sample points in the hand coordinate frame (\Secref{segm-sup}). We compare with 2 variants: uniformly sampling in the hand frame \& removing negative points from hand masks. We find our strategy to work the best (\Tableref{mask_sampling}).

\subsection{Visualizations}
We compare the mesh generated by our model and AC-SDF (trained on \obman - best baseline) on zero-shot generalization to \mow (\Figref{view_gen_viz}) and \core~\cite{Lomonaco2017CORL}(\Figref{view_gen_core}). For this, we sample points uniformly in a $64 \times 64 \times 64$ volume, predict their occupancies or SDF from the network and run marching cubes~\cite{lorensen1987marching}. We project the mesh into the input image \& render it in different views. Our model captures the visual hull of the object, as evidenced by the projection of the mesh onto the image, and generates more coherent shapes than AC-SDF, which often reconstructs disconnected and scattered shapes. More visualizations are in supplementary.

\newcommand\imgwidth{.1125}
\newcommand\x{.1}

\newcommand{\putvis}[6]{
\raisebox{#3}{\rotatebox{90}{\noindent \small #1}} & 
\includegraphics[width=\imgwidth\textwidth]{figures/obj_gen_mow/#6/0_0_#2/#4.png} & 
\includegraphics[width=\imgwidth\textwidth]{figures/obj_gen_mow/#6/0_0_#2/#5_0.png} &
\includegraphics[width=\imgwidth\textwidth]{figures/obj_gen_mow/#6/0_0_#2/#5_8.png} &
\includegraphics[width=\imgwidth\textwidth]{figures/obj_gen_mow/#6/0_0_#2/#5_12.png}
}
\begin{figure*}[t]
    \setlength{\tabcolsep}{1pt}
    \resizebox{\linewidth}{!}
    {
    \begin{tabular}[h]{p{0.3cm} c c c c p{0.3cm} c c c c}
        \putvis{GT}{197}{0.4cm}{image}{inp_obj}{ac_sdf} &
        \putvis{GT}{250}{0.4cm}{image}{inp_obj}{ac_sdf} \\
        \putvis{AC-SDF}{197}{0.1cm}{cam_mesh}{cHoi}{ac_sdf} &
        \putvis{AC-SDF}{250}{0.1cm}{cam_mesh}{cHoi}{ac_sdf} \\
        \putvis{\name}{197}{0.1cm}{cam_mesh}{cHoi}{ours} &
        \putvis{\name}{250}{0.1cm}{cam_mesh}{cHoi}{ours} \\ 
    \end{tabular}
    }
    \caption{{\bf Visualizations on \mow object generalization split.} We show the object mesh projected onto the image and rendered in different views for our HORSE model and compare with the AC-SDF model trained on \obman dataset with 3D supervision (best baseline model). We also show the ground truth (GT) object model. We observe that our model is able to predict the object shape more accurately than AC-SDF which often reconstructs smaller and disconnected shapes.}
    \figlabel{view_gen_viz}
\end{figure*}

\renewcommand{\putvis}[6]{
\raisebox{#3}{\rotatebox{90}{\noindent \small #1}} & 
\includegraphics[width=\imgwidth\textwidth]{figures/obj_gen_core/#6/0_0_#2/#4.png} & 
\includegraphics[width=\imgwidth\textwidth]{figures/obj_gen_core/#6/0_0_#2/#5_0.png} &
\includegraphics[width=\imgwidth\textwidth]{figures/obj_gen_core/#6/0_0_#2/#5_8.png} &
\includegraphics[width=\imgwidth\textwidth]{figures/obj_gen_core/#6/0_0_#2/#5_12.png}
}

\begin{figure*}[t]
    \setlength{\tabcolsep}{1pt}
    \resizebox{\linewidth}{!}
    {
    \begin{tabular}[h]{p{0.3cm} c c c c p{0.3cm} c c c c}
        \putvis{AC-SDF}{90}{0.1cm}{cam_mesh}{cHoi}{ac_sdf} &
        \putvis{AC-SDF}{220}{0.1cm}{cam_mesh}{cHoi}{ac_sdf} \\
        \putvis{\name}{90}{0.1cm}{cam_mesh}{cHoi}{ours} &
        \putvis{\name}{220}{0.1cm}{cam_mesh}{cHoi}{ours} \\
    \end{tabular}
    }
    \caption{{\bf Visualizations on zero-shot generalization to \core~\cite{Lomonaco2017CORL}.} We show the object mesh projected onto the image and rendered in different views on Core50. HORSE predicts better shapes than AC-SDF (best baseline, often leads to artifacts).
    } 
    \figlabel{view_gen_core}
\end{figure*}

\subsection{Limitations} 
\noindent {\bf Inaccurate hand pose.} We use predictions from FrankMocap for hand pose \& camera parameters. Note that the sampled points do not cover the entire object if the hand pose is not accurate, due to mis-projection into the image plane. This leads to exclusion of points in certain parts of the object (\Figref{inaccurate_pose}). %

\setlength{\intextsep}{0pt}%
\begin{wrapfigure}[6]{r}{0.45\textwidth}
    \centering
    \includegraphics[width=\linewidth]{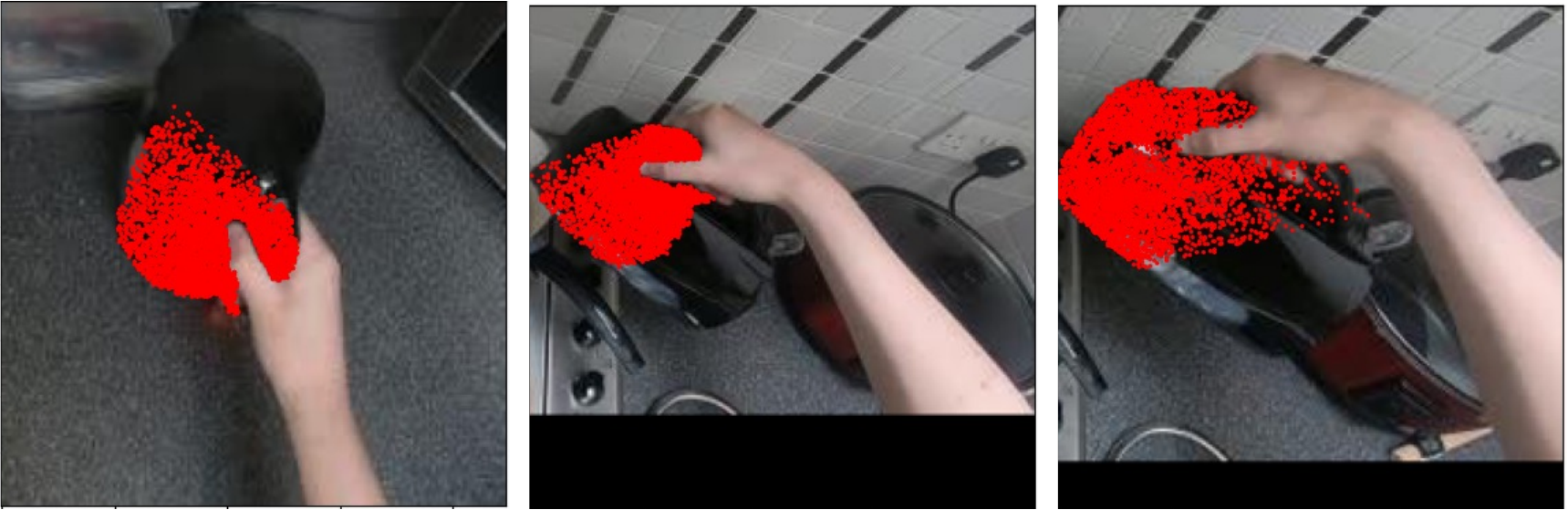}
    \caption{{\color{red}{Sampled}} points do not cover the entire object if hand pose is inaccurate.
    }
    \figlabel{inaccurate_pose}
\end{wrapfigure}
\noindent {\bf Limited object views.} Videos in the wild often do not capture 360$^{\circ}$ view of the object, \eg kettle in~\Figref{inaccurate_pose}. This is different than lab settings where the interactions are often constrained \& multi-camera setup is used to capture all sides of the object.

\section{Conclusion}
We present an approach for reconstructing hand-held objects in 3D from a single image. We propose modules to extract supervision from in-the-wild videos \& learn data-driven 3D shape priors from synthetic \obman to circumvent the need for direct 3D supervision. Experiments show that our approach generalizes better to novel objects in the wild than baselines trained using 3D supervision. Future directions include jointly optimizing the hand pose with the object shape to deal with inaccurate hand poses or incorporating additional cues, \eg contact priors.

\boldparagraph{Acknowledgements} We thank Ashish Kumar, Erin Zhang, Arjun Gupta, Shaowei Liu, Anand Bhattad, Pranay Thangeda \& Kashyap Chitta for feedback on the draft. This material is based upon work supported by NSF (IIS2007035), NASA (80NSSC21K1030), DARPA (Machine Common Sense program), an Amazon Research Award, an NVIDIA Academic Hardware Grant, and the NCSA Delta System (supported by NSF OCI 2005572 and the State of Illinois).

\bibliographystyle{splncs04}
\bibliography{biblioLong, refs}
\end{document}